\documentclass[letterpaper, 10 pt, conference]{ieeeconf}  % Comment this line out if you need a4paper

\IEEEoverridecommandlockouts                              % This command is only needed if 
\usepackage{graphicx}
\usepackage{epstopdf}
\usepackage{subcaption}
\usepackage{color}
\usepackage{multirow}
\usepackage[table,xcdraw]{xcolor}
\usepackage{caption}
\usepackage{array} 
\newcolumntype{L}[1]{>{\raggedright\let\newline\\\arraybackslash\hspace{0pt}}m{#1}} 
\newcolumntype{C}[1]{>{\centering\let\newline\\\arraybackslash\hspace{0pt}}m{#1}} 
\newcolumntype{R}[1]{>{\raggedleft\let\newline\\\arraybackslash\hspace{0pt}}m{#1}}
\usepackage{algorithm}
\usepackage{algpseudocode}

\usepackage{cite}
\usepackage{dirtytalk}
\usepackage[labelformat=simple]{subcaption}

\DeclareCaptionLabelFormat{subcaptionlabel}{\normalfont(\textbf{#2}\normalfont)}
\title{\LARGE \bf
LightFormer: An End-to-End Model for Intersection Right-of-Way Recognition Using Traffic Light Signals and an Attention Mechanism
}

\author{Zhenxing Ming, Julie Stephany Berrio, Mao Shan, Eduardo Nebot and Stewart Worrall%
\thanks{This work has been supported by the Australian Centre for Robotics (ACFR). The authors are with the ACFR at the University of Sydney (NSW, Australia). E-mails: zmin2675@uni.sydney.edu.au, \{j.berrio, m.shan, e.nebot, s.worrall\}@acfr.usyd.edu.au}%
}

\begin{document}

\maketitle
\thispagestyle{empty}
\pagestyle{empty}

%%%%%%%%%%%%%%%%%%%%%%%%%%%%%%%%%%%%%%%%%%%%%%%%%%%%%%%%%%%%%%%%%%%%%%%%%%%%%%%%
\begin{abstract}
For smart vehicles driving through signalised intersections, it is crucial to determine whether the vehicle has right-of-way given the state of the traffic lights. To address this issue, camera-based sensors can be used to determine whether the vehicle has permission to proceed straight, turn left or turn right. This paper proposes a novel end-to-end intersection right-of-way recognition model called LightFormer to generate right-of-way status for available driving directions in complex urban intersections. The model includes a spatial-temporal inner structure with an attention mechanism, which incorporates features from past image to contribute to the classification of the current frame's right-of-way status. In addition, a modified, multi-weight arcface loss is introduced to enhance the model's classification performance. Finally, the proposed LightFormer is trained and tested on two public traffic light datasets with manually augmented labels to demonstrate its effectiveness. The code is available for use at: https://github.com/DanielMing123/LightFormer
\end{abstract}

%\begin{IEEEkeywords}
%computer vision, transformer, attention mechanism.
%\end{IEEEkeywords}

\section{Introduction}\label{introduction}

Visual perception technology is commonly used to determine whether a vehicle has right-of-way in an urban intersection containing traffic lights based on the colour of the active light. As a result, traffic light detection and recognition techniques have significantly advanced in recent years. Existing work from both research and industry has demonstrated traffic light detection and recognition methods based on object detection. This approach consists of three components: 1) Locating each traffic light position in the image frame, 2) Making projections concerning each traffic light to acquire their 3D position, and 3) Binding each traffic light with each driving lane at the current scene based on their 3D position. 
This existing pipeline is suitable for processing traffic light scenes in basic road environments with few traffic lights and a straightforward road topology. In such situations, the post-processing algorithm can effectively establish the correlation between driving lanes and traffic lights. However, this approach may not work well for complex urban intersections. For instance, when two intersections are close together, multiple traffic lights and driving lanes for two different intersections may simultaneously exist in the same camera image, making it difficult to distinguish the correct signal. As a result, the third step of the pipeline may be error-prone and lead to poor accuracy in the classification of the right-of-way status.

\begin{figure}[t]
\centering
{\includegraphics[trim={0 4cm 0 2cm},clip,width=0.98\columnwidth]{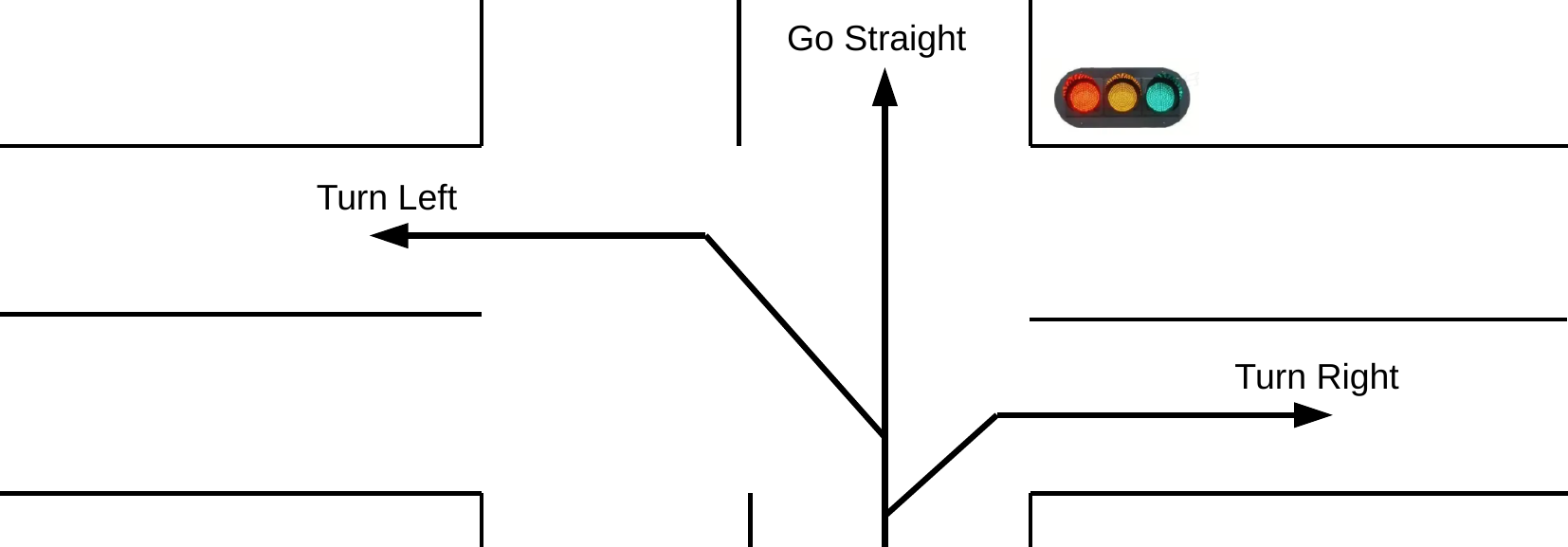}}
\caption{Available driving directions of a typical urban intersection considered in this paper}\label{inter}

\end{figure}

This paper proposes LightFormer, an end-to-end intersection right-of-way recognition algorithm. It takes consecutive frames containing traffic lights as the input, and outputs the right-of-way status for the corresponding driving directions at the intersection.
As shown in Fig. \ref{inter}, we assume three available driving directions exist for a lane at an intersection in a typical right-hand driving environment. This paper considers the possible turning options for most urban intersections: 1) Go straight, 2) Turn left, and 3) Turn right. For right-hand driving, a right turn generally poses lower risk compared to left turns as it does not cross oncoming traffic. Thus, for this work we consider that right turns can be made when the traffic light displays a green signal for going straight. Due to this, our algorithm focuses on the right-of-way status for the \textit{go straight} and \textit{turn left} directions. This results in four states in total for our model to predict: \textit{go straight pass}, \textit{go straight stop}, \textit{left turn pass}, and \textit{left turn stop}. The proposed model can be trivially applied to left-hand driving environments when trained with the corresponding data.

\begin{figure}[t]
\centering
{\includegraphics[trim={1.5cm 0cm 0cm 0cm},clip,width=0.98\columnwidth]{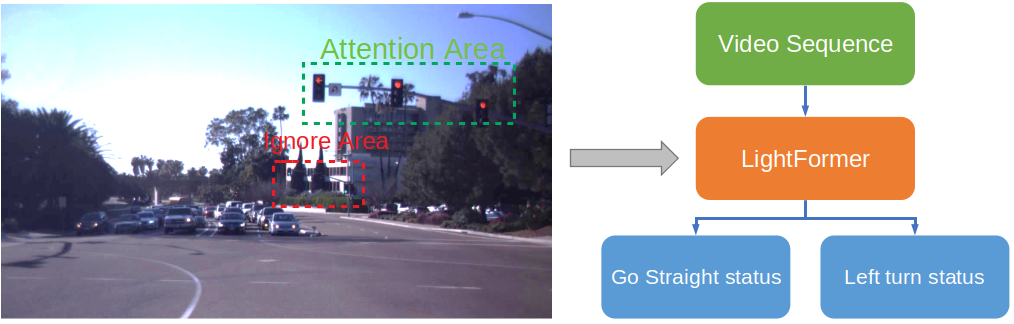}}
\caption{The new intersection right-of-way recognition pipeline based on the proposed algorithm}\label{pipe}
\end{figure}

The proposed new pipeline based on LightFormer is shown in Fig. \ref{pipe}. The end-to-end nature of this approach removes all computationally expensive post-processing from existing strategies and enables LightFormer to directly generate the right-of-way status for the available driving directions. The attention mechanism allows our model to determine which traffic light to pay attention to, and provide information on the right-of-way status for each driving direction.
The model includes a spatial-temporal inner structure that enables it to incorporate past image features and contribute to the current frame's prediction. Additionally, a modified arcface loss \cite{arcface} introduced here as `multi-weight arcface loss' is introduced to enhance the model's classification performance. Finally, we trained and tested the model on the Bosch \cite{Bosch} and Kaggle Lisa \cite{lisa1,lisa2} public traffic light datasets with manually augmented labels, and conducted an ablation study for each dataset to verify the feasibility of the proposed algorithm.

The main contributions of this paper are summarized below. 
\begin{itemize}
    \item We formulate a spatial-temporal inner structure, which enables the model to leverage the image features from the historical frames and contribute to the classification of the current frame. 
    \item A multi-weight arcface loss is introduced to enable our model to perform general clustering inside the neural network, improving our model's robustness and classification performance.
    \item The proposed LightFormer model is trained and tested on the public Bosch and Kaggle Lisa traffic light datasets with manually augmented labels to demonstrate the effectiveness of the proposed method.
\end{itemize}

The remainder of this paper is structured as follows: Section \ref{literature} provides an overview of related research and identifies the key differences between this study and previous publications. Section \ref{model} outlines the general framework of LightFormer and offers a detailed explanation of the implementation of each module. Section \ref{simulation} presents the findings of our experiments. Finally, Section \ref{conclusion} provides discussions and outlines potential future research directions.

\section{Related Work}\label{literature}
Following the conventional pipeline, researchers have proposed traffic light detection and recognition algorithms based on object detection techniques. For example, in \cite{rtld}, authors design a decision-tree-based \cite{dt} hand-crafted feature and add it to YOLO v3 \cite{yolo} to improve the algorithm robustness under different illumination conditions. Furthermore, in \cite{mdpiyolov4}, authors used a lightweight ShuffleNetv2 network \cite{shufflenetv2} to replace the original CSPDarkNet53 network of YOLOv4 \cite{yolov4} in order to increase the inference speed. A novel attention mechanism called $CS^{2}A$ is added to improve the feature extraction process, and the YOLOv4 algorithm's sensitivity to small objects is enhanced. Similarly, the authors of \cite{improvedyolov4} solve the problem of the YOLOv4 algorithm's insensitivity to small objects by creating a shallow feature enhancement, and bounding box uncertainty prediction mechanisms to improve the detection precision for traffic light detection and recognition. These algorithms mainly focus on improving the YOLO series inner structure to enhance the detection speed and precision of small objects like traffic lights. But they still rely on post-process algorithms to bind the traffic lights with each driving lane.

In contrast, some researchers have abandoned the technical solutions based on deep learning and turned to traditional image-processing methods for traffic light detection and recognition. The authors in \cite{tits} proposed an adaptive background suppression filter algorithm to highlight the traffic light candidate regions while suppressing the undesired background. In addition, a modified HOG feature detector \cite{hog} incorporating the local colour features was invented to do the traffic light recognition task. The algorithm is purely based on image processing techniques and achieved a decent real-time performance of about 15 fps under various illumination conditions. The authors of \cite{freq} used a high-speed camera and a frequency pattern recognition algorithm to detect traffic lights. The band-pass and Kalman filters were employed to extract useful information for detection. Both image-processing based algorithms have superior real-time performance when compared to deep-learning based algorithms, but due to heuristic design, they are not data-driven and are unable to self-evolve. 

To improve the robustness of the conventional pipeline, a few researchers also consider maps to strengthen the association between traffic lights and driving lanes. For example, the authors in \cite{lidar+hdmap} adopt a HD map  \cite{hdmap} and lidar data for traffic light detection and recognition. Due to the existence of HD maps, their algorithm can acquire the 3D position of the traffic lights in advance. In the meantime, the onboard lidar sensor can simultaneously verify the 3D position of the traffic light by measuring each static object’s distance. The dual insurance of HD map and lidar measurements ensures that the position of traffic lights at intersections can be detected very accurately. Then, based on the 3D position of each traffic light they cropped the image and fed it to YOLO to do traffic light detection and recognition.
Similarly, the authors of \cite{digitmap} adopt a digital map to determine the region-of-interest area in the image. Thus, the object detection algorithm's computational cost and false detection rate were significantly reduced. Traffic light detection and recognition based on object detection and the map shows promising performance but suffers from high implementation costs. Not all urban roads have HD maps. Once the vehicle travels to a country road without HD map support, such algorithms will fail.

\section{LightFormer}\label{model}

Combining multiple image features across historical frames can provide a robust and accurate classification result in the current frame. This study introduces a novel transformer-based framework for intersection right-of-way recognition tasks, which efficiently consolidates spatial-temporal features from historical image features utilizing an attention mechanism.

\subsection{Overall Architecture}

\begin{figure}[h]
\centering
\vspace{2mm}
\includegraphics[trim={1.5cm 0.5cm 1.5cm 0.5cm},clip, width=0.98\columnwidth] {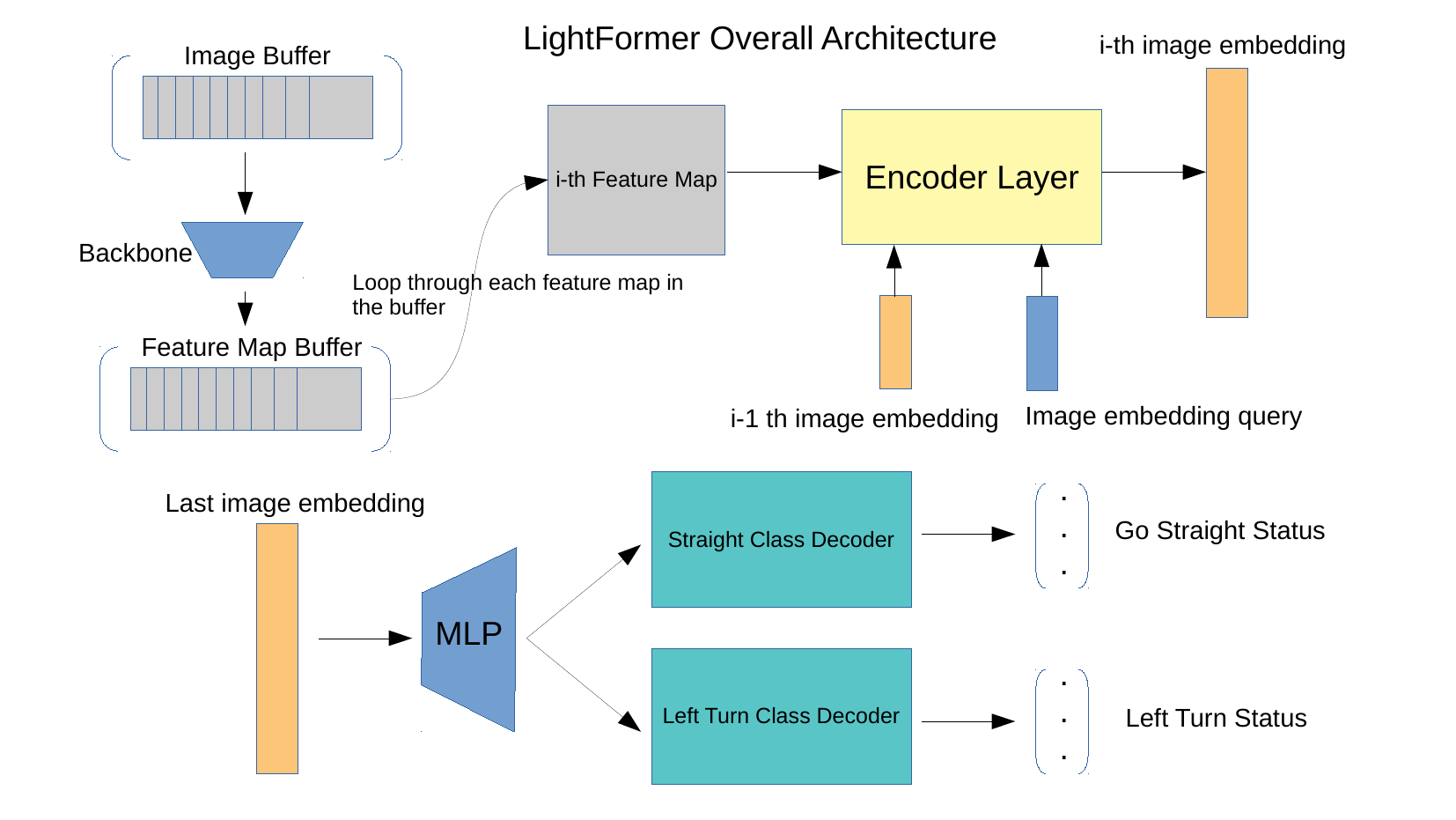}\caption{LightFormer overall architecture}\label{fig1}\end{figure}

\begin{figure}[t]
\vspace{2mm}
\centering
\includegraphics[trim={0 0 5cm 2cm},clip, width=0.98\columnwidth]{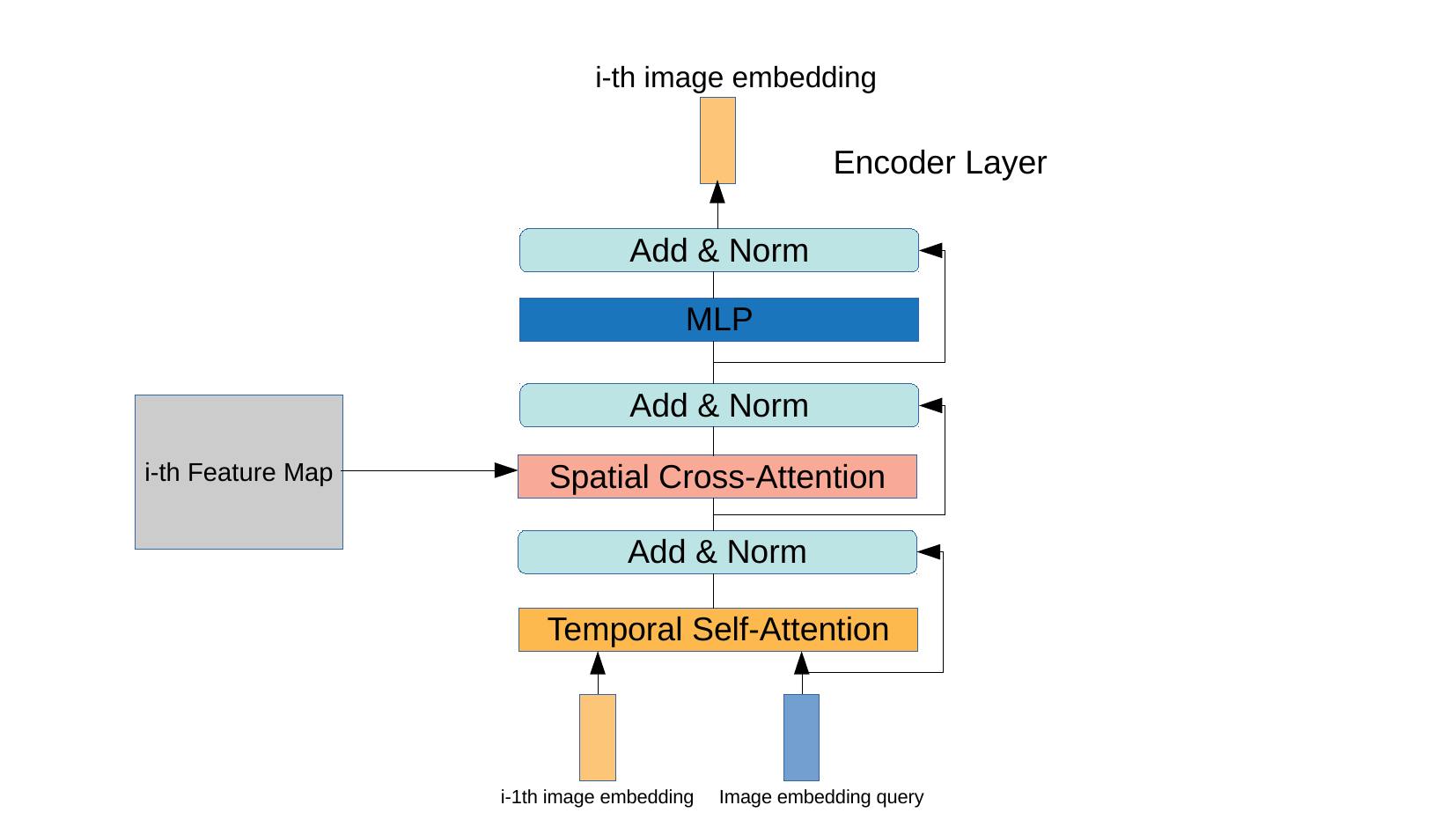}
\caption{Inner structure of the encoder layer}\label{fig2}
\end{figure}

As exhibited in Fig \ref{fig1},  our LightFormer model consists of three main modules: image buffer, encoder layer, and class decoder. The inner structure of the encoder layer is exhibited in Fig. \ref{fig2}.

In Fig. \ref{fig2}, the inner structure of the encoder layer follows the conventional structure of the transformer \cite{transformer}, except for the three customized designs, namely the image feature embedding query, temporal self-attention (TSA), and spatial cross-attention (SCA). The image feature embedding query is a trainable vector parameter comprising $D$ dimensions, designed to extract image feature embeddings from the historical and current frames utilizing the attention mechanism. Temporal self-attention and spatial cross-attention are two attention layers working with the image feature embedding query. These are used to look up the spatial image features from the current frame and aggregate the temporal features from the historical images feature embedding. The two class decoders in Fig. \ref{fig1} were implemented by the multi-weight arcface class decoder, a modified version of arcface that takes the last image embedding feature generated by the encoder layer as input, then conducts cluster analysis to filter out interference to compute the prediction.

We construct an image buffer containing $N$ images, the first $N-1$ images come from the historical frames, and the last image is the current frame. We feed the $N$ images to the backbone network, which is a ResNet18, and obtain $N$ image feature maps $\left \{ M^{i}  \right \} _{i=1}^{N} $, where $M^{i} $ is the feature map of the $i$-th image in the buffer. Then, we iteratively walk through each image feature map in the buffer and preserve each previous output image feature embedding $\left \{ E^{i}  \right \} _{i=1}^{N-1} $ as a history feature embedding. During each iteration, we use an image feature embedding query $Q$ to query the temporal information from the historical image feature embedding via the temporal self-attention layer. This aggregates the information from the past image frames, to inquire about the spatial image information from the image feature at the current $i$-th feature map via the spatial cross-attention layer. Finally, it outputs the image feature embedding vector, which aggregates the $N$ features in the buffer and feeds to the class decoder layer. A multi-weight arcface class decoder takes the embedding vector as input to cluster and predicts the right-of-way status of the driving direction at the intersection.

\subsection{Image Embedding Query}
We predefine a vector-like learnable parameter $Q\in R^{1\times D} $ as the image embedding query of the LightFormer, where D is the dimension of the query. This query aggregates the information from current and historical image frames and will be fed into the final class decoder.

\subsection{Temporal Self-Attention}
The inclusion of temporal information is essential for accurate predictions by the visual system. For instance, when judging the traffic light status there are occasional visual occlusions of the traffic light that can severely impact the model's predictive accuracy. To mitigate or even overcome this challenge, a temporal self-attention layer aggregates temporal information by incorporating the historical image embedding feature.

The temporal self-attention layer has two inputs: the embedding query Q at $i$-th feature map $M^{i}$ in the buffer and the historical image embedding vector $E^{i-1}$, generated after iterating over the $ i-1$-th feature map in the buffer. This temporal connection can be described as follows:
\begin{equation}\label{eq1}
    TSA\left ( Q,E^{i-1}  \right ) =Multihead(Q,K,V)
\end{equation}

The inner core of our temporal self-attention layer is the multi-head attention module, as in Equation \ref{eq1}. The $K$ and $V$ equal $E^{i-1}$, enabling the image embedding query $Q$ to aggregate the information from the past image embedding. Especially when operating on the first feature map $M^{1}$ in the buffer, since no image embedding $E^{0}$ exists, in this case, $K=V=Q$ and temporal self-attention $TSA\left ( Q,E^{i-1}  \right )$ degenerates to a simple self-attention layer $TSA\left ( Q,Q \right )$ without temporal information. 

\begin{figure*}[t]
\vspace{2mm}
     \centering
     \begin{subfigure}[]{0.32\textwidth}
         \centering
         \includegraphics[width=\textwidth]{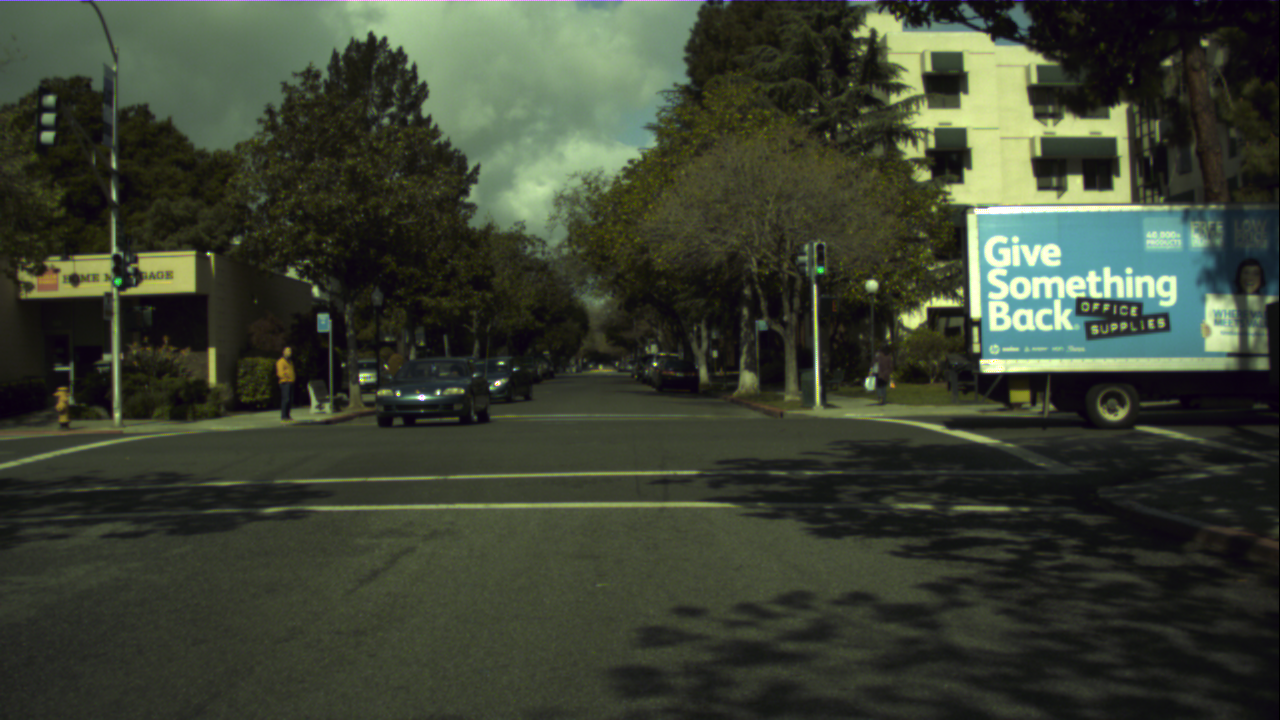}
         \caption{}
         \label{fig:y equals x}
     \end{subfigure}
     \begin{subfigure}[]{0.32\textwidth}
         \centering
         \includegraphics[trim={0 6cm 0 2.3cm},clip, width=\textwidth]{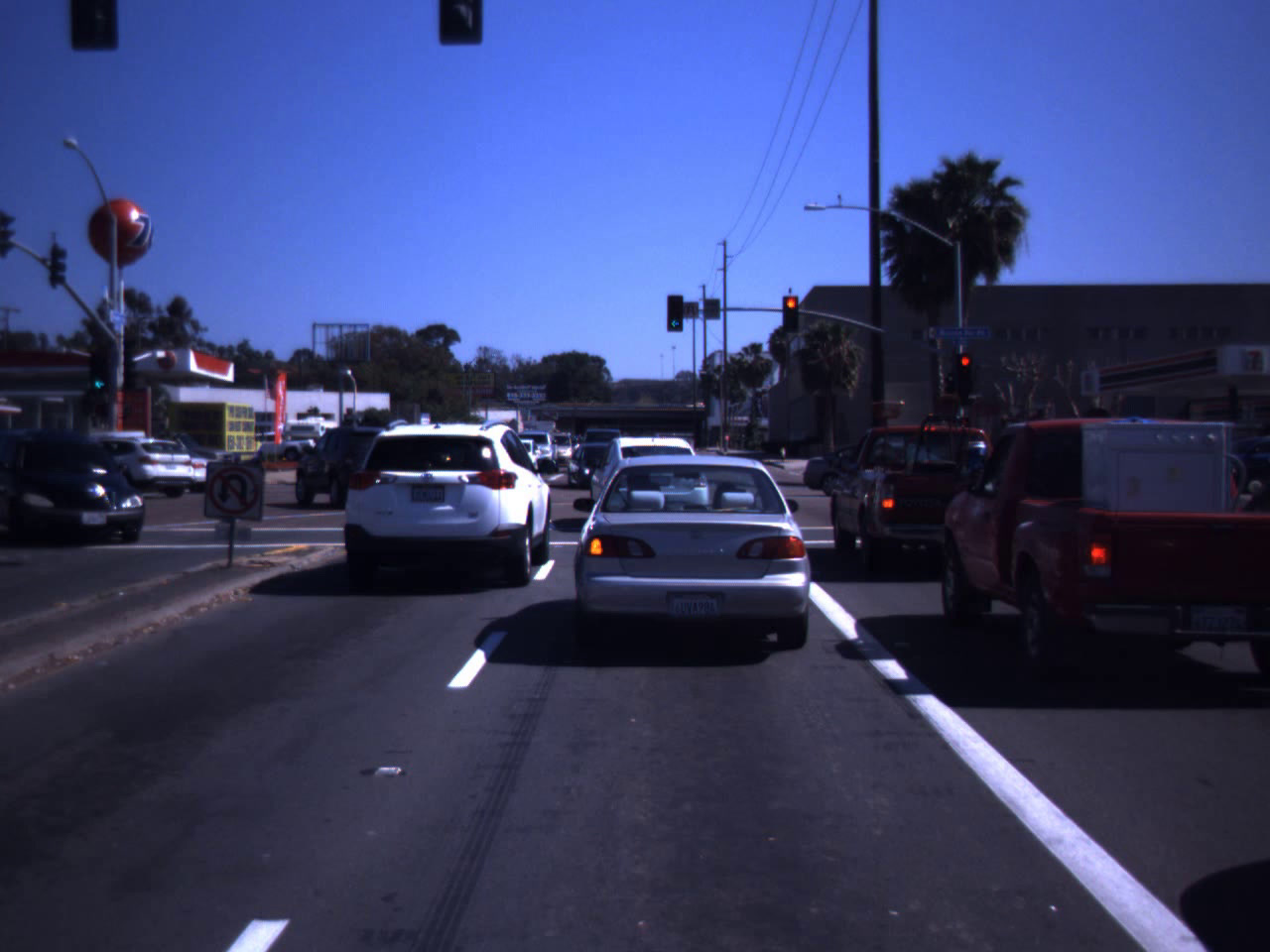}
         \caption{}
         \label{fig:three sin x}
     \end{subfigure}
     \begin{subfigure}[]{0.32\textwidth}
         \centering
         \includegraphics[trim={0 6cm 0 2.3cm},clip, width=
         \textwidth]{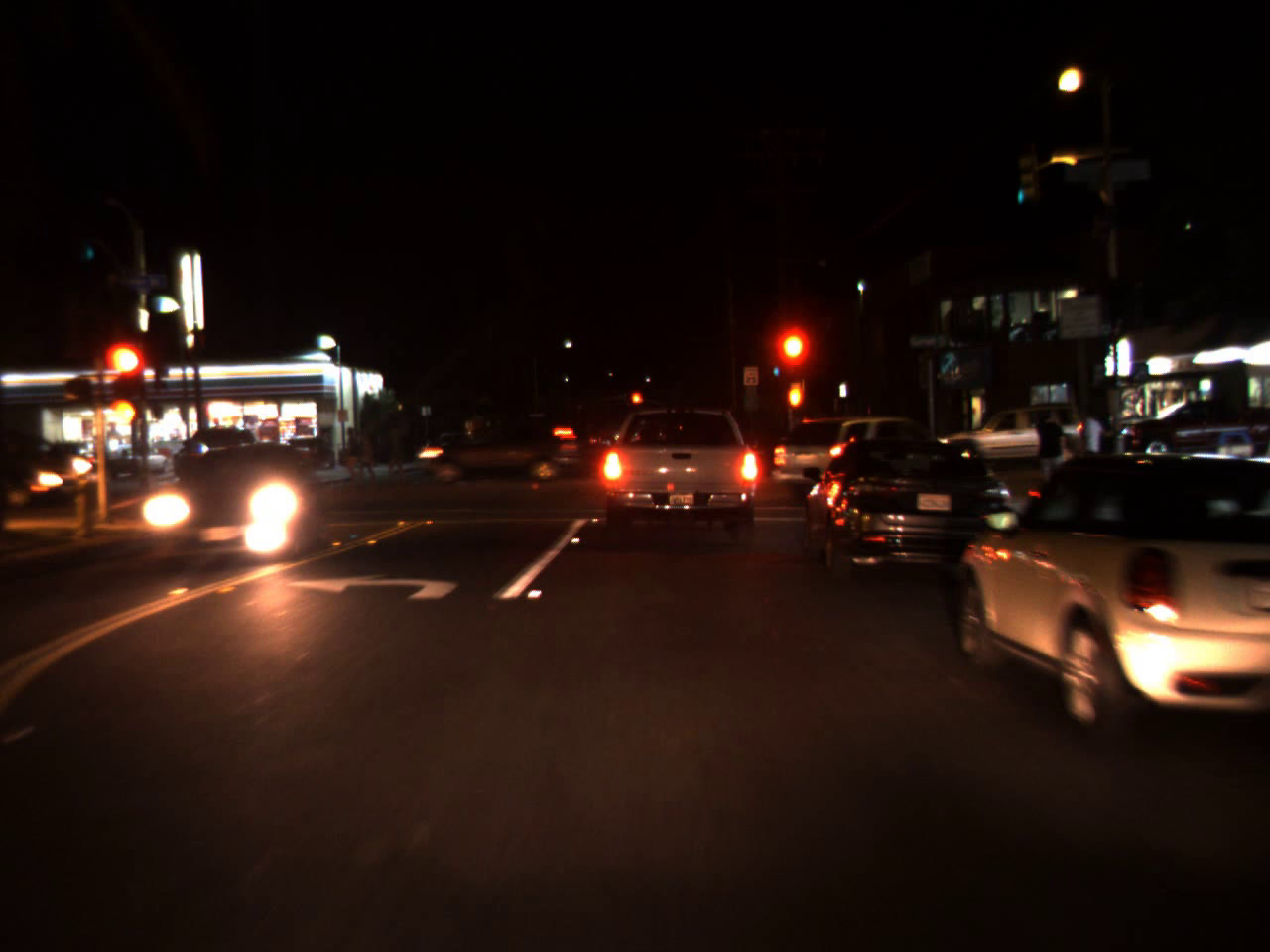}
         \caption{}
         \label{fig:five over x}
     \end{subfigure}
        \caption{\small Sample images for three different scenarios in the 
        Bosh and Lisa datasets. (a) Bosch day-time, (b) Lisa day-time, and (c) Lisa night-time.}
        \label{fig:three graphs}
\end{figure*}

Compared with the traditional stacking strategy, which directly stacks historical feature maps into a single large feature map to aggregate temporal information, our approach achieved an RNN-style attention mechanism inside the buffer. As a result, this structure can effectively model long temporal dependency and shows robustness towards noisy inputs.% suffer less noise information.
\subsection{Spatial Cross-Attention}
The spatial image information is the key for the visual system to understand the environment captured by the camera sensor. Some areas in the image may contain important information for our task, i.e., the currently lit traffic light colour. We have developed an attention-based spatial cross-attention layer to better acquire the critical information from the image feature map. Our spatial cross-attention layer is designed based on a deformable attention \cite{deformable} module, which is resource-efficient. Compared with the traditional multi-head attention module, the deformable attention module lets our image embedding query $Q$ only interact with specific areas of interest in the input feature map $M^{i}$. Thus, it can save significant computation power. The process of spatial cross-attention can be described as follow:
\begin{equation}
    SCA(Q,M_{p}^{i} ) = \sum_{V\in\left \{ Q,M_{p}^{i}  \right \}  } DeformAttn\left ( Q,M_{p}^{i},V   \right ) 
\end{equation}
where $Q$, in this case, is the output of the temporal self-attention layer, $M_{p}^{i} $ is the feature vector on the feature map $M^{i}$ at location $p=(u,v)$.

\subsection{Multi-Weight Arcface Class Decoder}
The arcface class decoder, designed initially to assist the deep face recognition task, can produce a safety margin and provide clear borders between different classes. Moreover, although many different shapes of traffic lights exist, i.e., circle light and arrow light, their semantic information is very similar. Thus, we developed a modified version of the arcface class decoder, the multi-weight arcface class decoder, to enable the model to do cluster analysis. Our model only classifies four categories, \textit{go straight pass}, \textit{go straight stop}, \textit{left turn pass}, and \textit{left turn stop}. There are four classes in total, and the underlying principle of the multi-weight arcface is incorporating $w$ cluster centres for each category. This results in $4\times w$ cluster centres in total. We apply the max function for each class of $w$ cluster centres to extract the least discriminative vector on behalf of our model's predicted probability and output to the cross-entropy loss function. This design would force our model to do a clustering operation based on the image embedding vector extracted from the encoder layer and optimise the image feature embedding vector extraction process. 

\section{Experimental Results}\label{simulation}
\subsection{Dataset}
The experiment is conducted on two public datasets, namely the Bosch small traffic light dataset and the Kaggle Lisa traffic light dataset. The Bosch dataset only contains day time scenarios, and the road environment is simple. In contrast, the Lisa dataset is much more challenging. It contains both day and night time scenarios. Most day time scenes are congested intersection scenes with lots of light interference. On the other hand, the night time scenes lack rich texture information and contain interference from street lights. The sample images for three scenarios are exhibited in Fig. \ref{fig:three graphs}.

\begin{figure}[H]
\begin{center}
{\includegraphics[width=0.95\columnwidth]{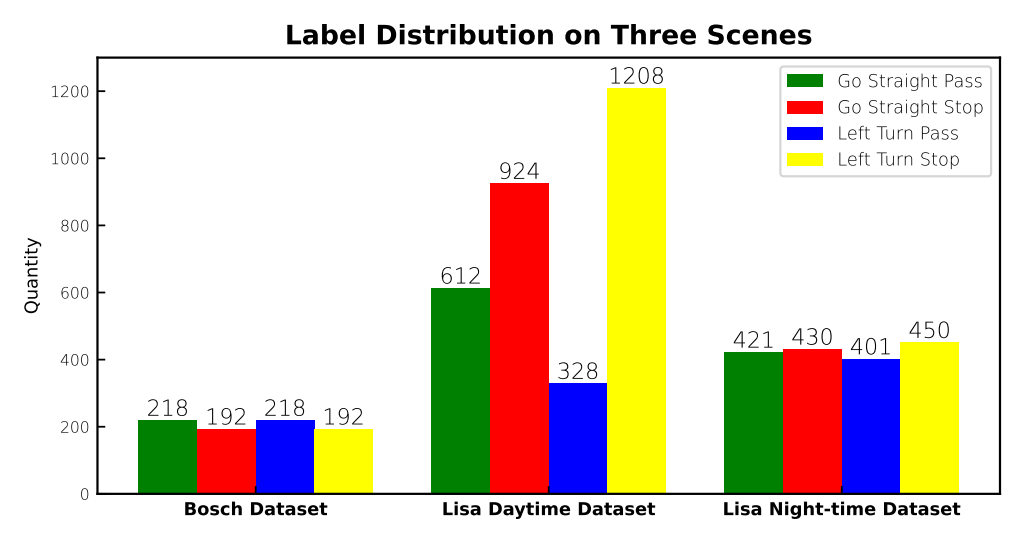}} %0.9
\caption{\small Label distribution on three datasets}\label{label_dist}
\end{center}
\end{figure}

Since these two existing public datasets were generated for the object detection task, and our model is designed for end-to-end intersection right-of-way status recognition for available driving directions, it is necessary to process and modify the labels of these datasets to adapt to our model. Our data label consists of four states: \textit{go straight pass}, \textit{go straight stop}, \textit{left turn pass}, and \textit{left turn stop}, which is consistent with the city intersection model exhibited in Fig. \ref{inter}. We manually label these four states for each frame of the two public datasets. For each traffic light, we determine whether the green light means pass, and yellow or red light means stop for each direction of travel. Each training and testing sample contains $N$ frames, and the last frame's traffic light signal determines the right-of-way status. During our processing of these two public datasets, the original data labels of each intersection scene have been updated accordingly. As a result, the distribution of the four states in each dataset is shown in Fig. \ref{label_dist}.

\subsection{Performance Evaluate Metrics}
In our experiment, we choose four metrics to evaluate each output status of our model. Those four metrics are accuracy, precision, recall, and F1-score. Accuracy is the primary evaluation metric describing the number of correct predictions over all predictions.
% and it can be calculated by equation \ref{eq3}. 
% \begin{equation}\label{eq3}
%     \textit{Accuracy}=\frac{TP+TN}{TP+TN+FP+FN}
% \end{equation}
% where TP refers to true positive, TN refers to true negative, FP refers to false positive, and FN refers to a false negative.
Precision is a metric used to measure how many positive predictions are correct.
% and it can be calculated by equation \ref{eq4}. 
% \begin{equation}\label{eq4}
%     \textit{Precision}=\frac{TP}{TP+FP}
% \end{equation}
The recall is another evaluation metric used to measure how many positive cases the classifier correctly predicted over all the positive cases in the data. It is sometimes also referred to as sensitivity. 
% Its calculation process is shown in equation  \ref{eq5}. 
% \begin{equation}\label{eq5}
%    \textit{Recall}=\frac{TP}{TP+FN}
% \end{equation}
Finally, F1-score is an evaluation metric combining both precision and recall. It is generally described as the harmonic mean of the two, balancing the two ratios (precision and recall), requiring both to have a higher value for the F1-score value to rise. 
% Its detailed calculation process is shown in equation \ref{eq6}
% \begin{equation}\label{eq6}
%     \textit{F1-score}=2\times \frac{Precision\times Recall}{Precision+Recall}
% \end{equation}


\begin{table*}[t]
\centering
\vspace{2mm}
\caption{Performance on Bosch dataset}\label{table1} 
\begin{tabular}{|l|cc|cc|cc|cc|}
\hline
\multicolumn{1}{|c|}{\cellcolor[HTML]{FFFFFF}}                                           & \multicolumn{2}{c|}{\textbf{Precision}}                       & \multicolumn{2}{c|}{\textbf{Recall}}                          & \multicolumn{2}{c|}{\textbf{F1-Score}}                        & \multicolumn{2}{c|}{\textbf{Accuracy}}                        \\ \cline{2-9} 
\multicolumn{1}{|c|}{\multirow{-2}{*}{\cellcolor[HTML]{FFFFFF}\textbf{R-o-w Status}}} & \multicolumn{1}{c|}{\textbf{Without TSA}} & \textbf{With TSA} & \multicolumn{1}{c|}{\textbf{Without TSA}} & \textbf{With TSA} & \multicolumn{1}{c|}{\textbf{Without TSA}} & \textbf{With TSA} & \multicolumn{1}{c|}{\textbf{Without TSA}} & \textbf{With TSA} \\ \hline
Go Straight Pass                                                                         & \multicolumn{1}{c|}{96.81\%}              & 100.00\%          & \multicolumn{1}{c|}{98.91\%}              & 100.00\%          & \multicolumn{1}{c|}{97.85\%}              & 100.00\%          & \multicolumn{1}{c|}{98.10\%}              & 100.00\%          \\ \hline
Go Straight Stop                                                                         & \multicolumn{1}{c|}{99.14\%}              & 100.00\%          & \multicolumn{1}{c|}{97.46\%}              & 100.00\%          & \multicolumn{1}{c|}{98.29\%}              & 100.00\%          & \multicolumn{1}{c|}{98.10\%}              & 100.00\%          \\ \hline
Left Turn Pass                                                                           & \multicolumn{1}{c|}{97.85\%}              & 100.00\%          & \multicolumn{1}{c|}{98.91\%}              & 98.91\%           & \multicolumn{1}{c|}{98.38\%}              & 99.45\%           & \multicolumn{1}{c|}{98.57\%}              & 99.52\%           \\ \hline
Left Turn Stop                                                                           & \multicolumn{1}{c|}{99.15\%}              & 99.16\%           & \multicolumn{1}{c|}{98.31\%}              & 100.00\%          & \multicolumn{1}{c|}{98.72\%}              & 99.58\%           & \multicolumn{1}{c|}{98.57\%}              & 99.52\%           \\ \hline
\end{tabular}
\end{table*}

\begin{table*}[]
\centering
\caption{Performance on Lisa day-time data}\label{table3}
\begin{tabular}{|l|cc|cc|cc|cc|}
\hline
\multicolumn{1}{|c|}{\cellcolor[HTML]{FFFFFF}}                                           & \multicolumn{2}{c|}{\textbf{Precision}}                       & \multicolumn{2}{c|}{\textbf{Recall}}                          & \multicolumn{2}{c|}{\textbf{F1-Score}}                        & \multicolumn{2}{c|}{\textbf{Accuracy}}                        \\ \cline{2-9} 
\multicolumn{1}{|c|}{\multirow{-2}{*}{\cellcolor[HTML]{FFFFFF}\textbf{R-o-w Status}}} & \multicolumn{1}{c|}{\textbf{Without TSA}} & \textbf{With TSA} & \multicolumn{1}{c|}{\textbf{Without TSA}} & \textbf{With TSA} & \multicolumn{1}{c|}{\textbf{Without TSA}} & \textbf{With TSA} & \multicolumn{1}{c|}{\textbf{Without TSA}} & \textbf{With TSA} \\ \hline
Go Straight Pass                                                                         & \multicolumn{1}{c|}{98.04\%}              & 100.00\%          & \multicolumn{1}{c|}{99.01\%}              & 99.01\%           & \multicolumn{1}{c|}{98.52\%}              & 99.50\%           & \multicolumn{1}{c|}{98.51\%}              & 99.50\%           \\ \hline
Go Straight Stop                                                                         & \multicolumn{1}{c|}{99.00\%}              & 99.02\%           & \multicolumn{1}{c|}{98.02\%}              & 100.00\%          & \multicolumn{1}{c|}{98.51\%}              & 99.51\%           & \multicolumn{1}{c|}{98.51\%}              & 99.50\%           \\ \hline
Left Turn Pass                                                                           & \multicolumn{1}{c|}{71.43\%}              & 96.94\%           & \multicolumn{1}{c|}{51.89\%}              & 89.62\%           & \multicolumn{1}{c|}{60.11\%}              & 93.14\%           & \multicolumn{1}{c|}{63.86\%}              & 93.07\%           \\ \hline
Left Turn Stop                                                                           & \multicolumn{1}{c|}{59.20\%}              & 89.42\%           & \multicolumn{1}{c|}{77.08\%}              & 96.88\%           & \multicolumn{1}{c|}{66.97\%}              & 93.00\%           & \multicolumn{1}{c|}{63.86\%}              & 93.07\%           \\ \hline
\end{tabular}
\end{table*}

\subsection{Experimental Settings}
In our experiment, we choose ResNet18 as our backbone to extract features from each image in the image buffer. The size of the image buffer $N$ is set to 10 and the number of cluster centres $w$ is set to 1. We conduct our experiment on an RTX3080 GPU with 10 GB memory on a Ubuntu 18 computer. On the Bosch dataset, the image sequences in the image buffer are two frames apart from each other. On the Lisa dataset, the image sequences in the image buffer are one frame apart from each other. Each image in two datasets is resized to $512\times960$ to adapt to our model. We choose Adam Optimizer with a 1e-4 initial learning rate and train our model for 15 epochs.

\subsection{Model Performance Analysis}
\subsubsection{Performance on Bosch Dataset}
The baseline model is the LightFormer model introduced in this paper. To demonstrate the importance of the temporal self-attention (TSA) module, we have conducted an ablation study by removing it from the base LightFormer model. Both models were trained and tested on the Bosch dataset, and their performance is exhibited in Table \ref{table1}.

%\begin{table}[H]
%\begin{center}
%\caption{Performance on Bosch Dataset}\label{table1} 
%  \begin{tabular}{ C{1.4cm}|L{1.2cm} | L{1.2cm} | L{1.2cm} | L{1.2cm}}
%    \hline
%    Right-of-Way Status & Precision & Recall & F1-Score  & Accuracy \\\hline 

%    Go Straight Pass & 100.00\% & 100.00\% & 100.00\% & 100.00\% \\ \hline
%    Go Straight Stop & 100.00\% & 100.00\% & 100.00\% & 100.00\% \\ \hline
%    Left Turn Pass & 100.00\% & 98.91\% & 99.45\% & 99.52\% \\ \hline
%    Left Turn Stop & 99.16\% & 100.00\% & 99.58\% & 99.52\% \\ \hline
%    \end{tabular}
%\end{center}
%\end{table}

% In Table \ref{table1}, our LightFormer model (with TSA) achieves 100\% in accuracy, precision, recall and f1-score for the go straight direction on the Bosch dataset. The performance on the left turn direction is slightly worse than the go straight direction, but it exceeds 99\% in all four metrics.  The LightFormer model (without TSA), compared with the baseline, the index of the four states in Table \ref{table1} is slightly lower on all metrics. This ablation study conducted on the Bosch dataset shows that the TSA module indeed improves the model's performance in judging the Right-of-Way status of two available driving direction. However, due to the small amount of data in the Bosch dataset, the performance difference between the two is not apparent on the test set.

In Table \ref{table1}, our LightFormer model (with TSA) achieves 100\% in accuracy, precision, recall and F1-score for the go straight direction on the Bosch dataset. The performance on the left turn direction is slightly worse than the go straight direction, but it exceeds 99\% in all four metrics. For the LightFormer model (without TSA), compared with the baseline, the index of the four states in Table \ref{table1} is slightly lower on all metrics. The experiment result shows that the TSA module improves the model's performance in judging the right-of-way status of two available driving directions by leveraging the temporal information. However, due to the small amount of data in the Bosch dataset, the performance difference between the two is not apparent on the test set.

\subsubsection{Performance on Lisa Dataset}
The Lisa dataset contains both daytime and night-time scenes. For the end-to-end model, the night-time scene is significantly more challenging than the daytime scene because many textures in the image are lost in the night-time scene. Our model was trained and tested with daytime and night-time data separately to verify the role of the temporal self-attention module and the multi-weight arcface class decoder module in improving model performance. Again, we chose LightFormer as our baseline and did an ablation study on the Kaggle Lisa daytime dataset by removing the TSA module. The performance of two models on the Kaggle Lisa daytime dataset is exhibited in Table \ref{table3}.
%\begin{table}[H]
%\begin{center}
%\caption{Performance on Lisa Day-time Data}\label{table3} 
%  \begin{tabular}{ C{1.4cm}|L{1.2cm} | L{1.2cm} | L{1.2cm} | L{1.2cm}}
%    \hline
%    Right-of-Way Status & Precision & Recall & F1-Score  & Accuracy \\\hline 
%    Go Straight Pass & 100.00\% & 99.01\% & 99.50\% & 99.50\% \\ \hline
%    Go Straight Stop & 99.02\% & 100.00\% & 99.51\% & 99.50\% \\ \hline
%    Left Turn Pass & 96.94\% & 89.62\% & 93.14\% & 93.07\% \\ \hline
%    Left Turn Stop & 89.42\% & 96.88\% & 93.00\% & 93.07\% \\ \hline
%    \end{tabular}
%\end{center}
%\end{table}
In Table \ref{table3}, the LightFormer baseline model still performs well in judging the right-of-way status for the available driving direction. Nevertheless, for left turn direction right-of-way status judgement, it performs slightly worse. This is because the left turn indicator appears in the form of a circle light at some intersections, and it appears as a left turn arrow light at other intersections. Besides, the left turn indicator is not as sharp as the straight indicator. Thus, it is more difficult to identify the left turn direction right-of-way status in the Kaggle Lisa dataset. Furthermore, the label distribution of the left turn direction is less balanced than the go straight direction, and we did not apply any data augmentation techniques, which also challenges our model to predict the right-of-way status of the left turn direction.
\begin{table*}[]
\centering
% \vspace{2mm}
\centering
\caption{Performance on Lisa night-time data with different $w$}\label{table5} 
\begin{tabular}{|l|cc|cc|cc|cc|}
\hline
\multicolumn{1}{|c|}{\cellcolor[HTML]{FFFFFF}}                                           & \multicolumn{2}{c|}{\textbf{Precision}}           & \multicolumn{2}{c|}{\textbf{Recall}}              & \multicolumn{2}{c|}{\textbf{F1-Score}}            & \multicolumn{2}{c|}{\textbf{Accuracy}}            \\ \cline{2-9} 
\multicolumn{1}{|c|}{\multirow{-2}{*}{\cellcolor[HTML]{FFFFFF}\textbf{R-o-w Status}}} & \multicolumn{1}{c|}{\textbf{w=1}} & \textbf{w=20} & \multicolumn{1}{c|}{\textbf{w=1}} & \textbf{w=20} & \multicolumn{1}{c|}{\textbf{w=1}} & \textbf{w=20} & \multicolumn{1}{c|}{\textbf{w=1}} & \textbf{w=20} \\ \hline
Go Straight Pass                                                                         & \multicolumn{1}{c|}{48.03\%}      & 94.44\%       & \multicolumn{1}{c|}{44.85\%}      & 100.00\%      & \multicolumn{1}{c|}{46.39\%}      & 97.14\%       & \multicolumn{1}{c|}{45.14\%}      & 96.89\%       \\ \hline
Go Straight Stop                                                                         & \multicolumn{1}{c|}{42.31\%}      & 100.00\%      & \multicolumn{1}{c|}{45.45\%}      & 93.39\%       & \multicolumn{1}{c|}{43.82\%}      & 96.58\%       & \multicolumn{1}{c|}{45.14\%}      & 96.89\%       \\ \hline
Left Turn Pass                                                                           & \multicolumn{1}{c|}{54.55\%}      & 94.44\%       & \multicolumn{1}{c|}{39.71\%}      & 100.00\%      & \multicolumn{1}{c|}{45.96\%}      & 97.14\%       & \multicolumn{1}{c|}{50.58\%}      & 96.89\%       \\ \hline
Left Turn Stop                                                                           & \multicolumn{1}{c|}{48.10\%}      & 100.00\%      & \multicolumn{1}{c|}{62.81\%}      & 93.39\%       & \multicolumn{1}{c|}{54.48\%}      & 96.58\%       & \multicolumn{1}{c|}{50.58\%}      & 96.89\%       \\ \hline
\end{tabular}
\end{table*}
Compared to the index of the baseline model in Table \ref{table3}, the model without the TSA module degrades on the go straight direction right-of-way status judgement and performs poorly on the left turn right-of-way status judgement. Due to the unbalanced label distribution in the left turn direction and lack of temporal information, the model without the TSA module only achieves 60.11\% for the \textit{left turn pass} status judgment and 66.97\% for the \textit{left turn stop} status judgment under the F1-score metric. In contrast, our baseline model achieves 93.14\% and 93\% performance under the F1-score metric, respectively. This experiment result reveals the fact that with the assistance of temporal information, the model can make a prediction with higher confidence and accuracy, especially when the model is dealing with challenging data or an unbalanced data label distribution. Thus, the TSA module plays a significant role in the LightFormer. Besides, the amount of data in the Lisa dataset is more than five times that of the Bosch dataset. Therefore, the test results on the Lisa dataset can better reflect the importance of the TSA module as well. Next, we will analyze and compare the performance of our model under the data training of the Lisa night scene in Table \ref{table5} and verify that the multi-weight arcface class decoder module can also significantly improve the model's performance.
The night-time data in the Lisa dataset is much more complex than the daytime data due to the lack of rich texture in the image. In this case, the outline of the lightbox of the traffic light is no longer visible, and the colour of the traffic light becomes the only clue for the available driving direction right-of-way status judgement. In addition, street lights in the night scenes also interfere with the model's judgment based on the light colour. As a result, even with the assistance of temporal information provided by the TSA module, our model performs poorly in the night scene, as exhibited in Table \ref{table5}. 
From Table \ref{table5}, we can see that the baseline model only achieves 46.39\% for the \textit{go straight pass} status judgment, 43.82\% for the \textit{go straight stop} status judgment, 45.96\% for the \textit{left turn pass} status judgment and 54.48\% for the \textit{left turn stop} status judgment under the F1-score evaluation metric. 
The performance is even lower than the random guess, in which you at least have a 50\% chance to figure out the correct status for these four states. Hence, only one cluster centre for each class can no longer provide correct guidance, and a cluster analysis mechanism, which takes the input image embedding feature and casts it on multiple cluster centres with the same category, is needed to filter out the interference features. In the experiment, the $w$ setting larger than 5 can achieve around 90\% accuracy for all four states, a very decent performance, and there is no significant change in performance if $w$ continues to increase. Thus, we set cluster centre $w$ to 20 as the redundant setting, which already reached the peak performance of the module, to introduce 20 cluster centres for each class in each decoder module. 
As exhibited in Table \ref{table5}, by increasing the number of cluster centres, the index in Table \ref{table5} for the modified model is boosted. As a result, the model achieves 97.14\% for the \textit{go straight pass} status judgement, 96.58\% for the \textit{go straight stop} status judgement, 97.14\% for the \textit{left turn pass} status judgement and 96.58\% for the \textit{left turn stop} status judgement under the F1-score evaluation metric. Compared with the baseline model, the index of the four states have been improved by at least 50\%. This experimental result verifies that the multi-weight arcface class decoder module positively improves the model's performance through cluster analysis. By optimizing the least discriminative vector, which on behalf of our model's predicted probability, our model is forced to perform cluster analysis based on the extracted image embedding vector extracted from the encoder layer. Through the clustering operation, the interference features in the image embedding vector are removed, facilitating our model to generate a better prediction output.

\subsection{Failure Case Analysis}
Although our algorithm achieved decent performance on both public datasets, there still exist several city intersection scenarios that our algorithm failed to generate the correct right-of-way status. Therefore, we conduct a failure case analysis on daytime and night-time scenes to identify the shortcomings of the proposed algorithm.
\subsubsection{Day-Time Failure Case Analysis}
\begin{figure}[H]
\centering
% \begin{subfigure}[]{0.23\textwidth}{\includegraphics[width=\textwidth]{Figs/block1.png}
% \caption{}\label{a}}
% \end{subfigure}
% \begin{subfigure}[]{0.23\textwidth}
%     \includegraphics[width=\textwidth]{Figs/block2.png}
%         \caption{}\label{b}
%     \end{subfigure}

    \begin{subfigure}[]{0.23\textwidth}
        \includegraphics[width=\textwidth]{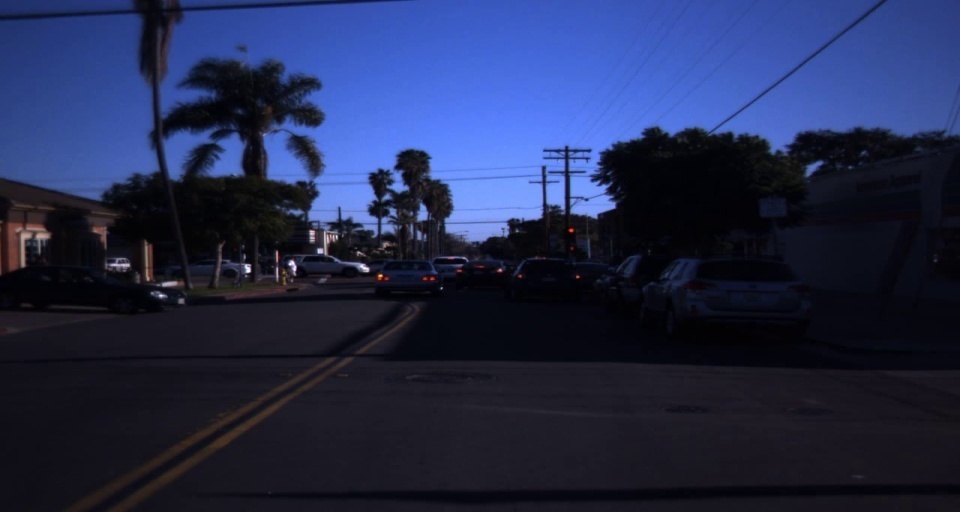}
        \caption{}\label{a}
    \end{subfigure}
    \begin{subfigure}[]{0.23\textwidth}
        \includegraphics[width=\textwidth]{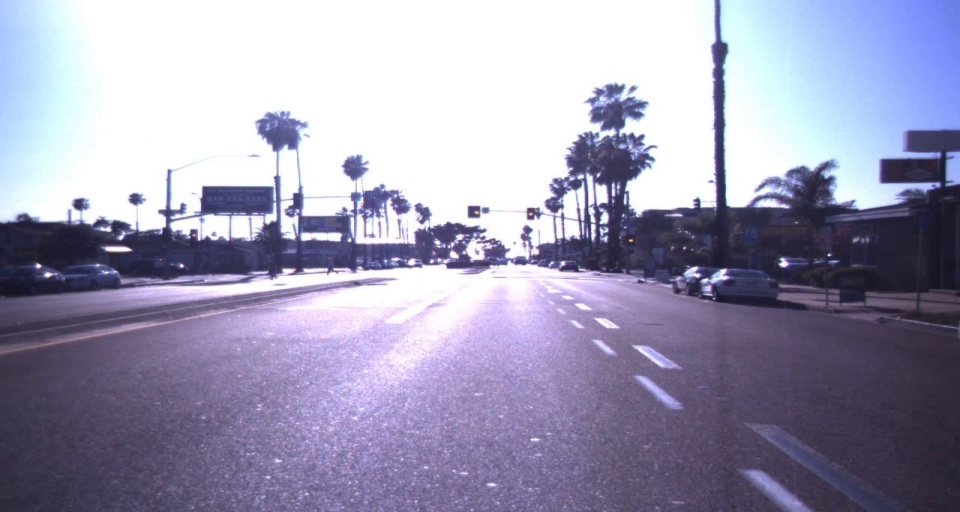}
        \caption{}\label{b}
    \end{subfigure}
    \caption{\small Day-time failure cases from two public datasets. (\textbf{a}) Dim light with traffic light far away from ego vehicle. Predict: \textit{go straight stop }, \textit{left turn pass}. Ground Truth (GT): \textit{go straight stop}, \textit{left turn stop}. (\textbf{b}) Bright light with traffic light far away from ego vehicle. Predict: \textit{go straight pass}, \textit{left turn pass}. GT: \textit{go straight stop}, \textit{left turn stop}.}\label{fig:6}
\end{figure}

We collected few typical day-time failure case scenes from two public datasets and exhibited them in Fig \ref{fig:6}. 
From Fig. \ref{a} and Fig. \ref{b} we discover that dramatic changes to the ambient light will affect the proposed algorithm's performance. If the ambient light is too dim or bright, and the ego vehicle is far away from the traffic light, the signal of the traffic light is too weak to be picked up by our model due to the severe interference from the ambient light environment. As a result, our proposed algorithm will fail to extract useful information and make an incorrect judgment.

\subsubsection{Night-Time Failure Case Analysis}
\begin{figure}[]
\vspace{2mm}
    \centering
   {\captionsetup{position=bottom,justification=centering}
       \begin{subfigure}[]{0.23\textwidth}{ \includegraphics[trim={0 8.3cm 0 0},clip,width=\textwidth]{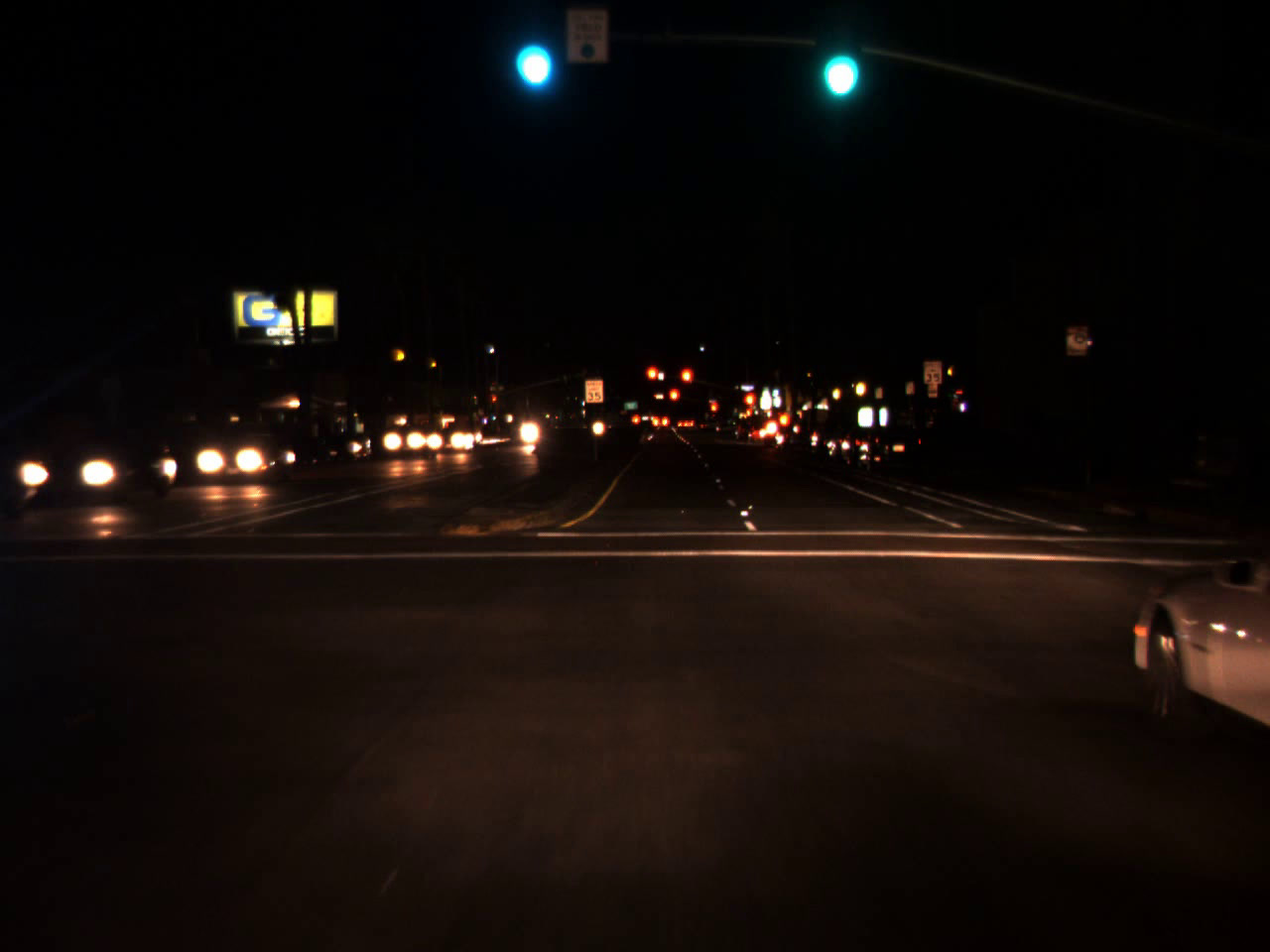}
        \caption{}\label{7a}}
    \end{subfigure}
    \begin{subfigure}[]{0.23\textwidth}
        \includegraphics[trim={0 6cm 0 2.3cm},clip, width=\textwidth]{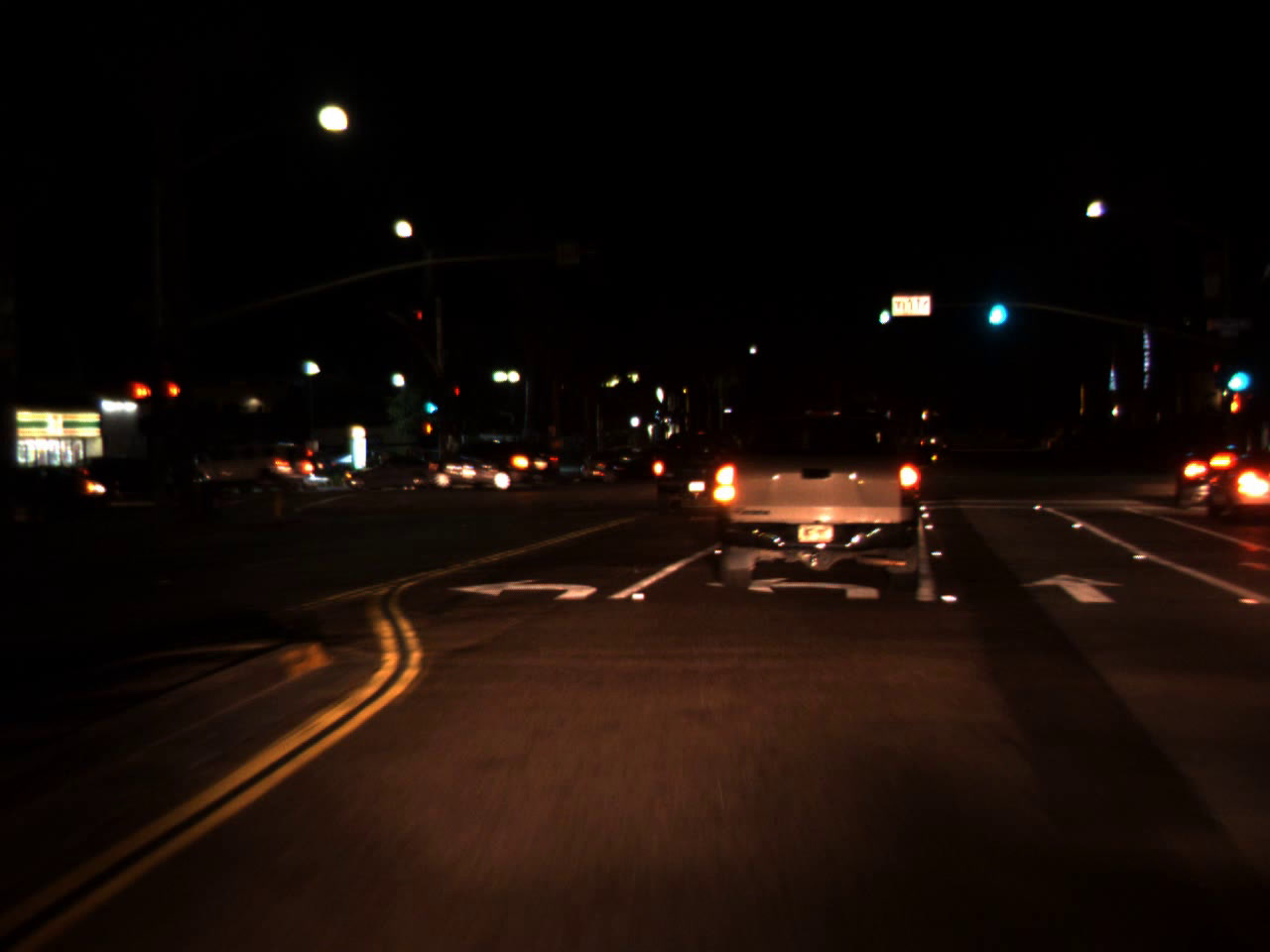}
        \caption{}\label{7b}
    \end{subfigure}
   }
    \caption{\small Night-time failure cases from Kaggle Lisa datasets. (\textbf{a}) Multiple light sources at night. Predict: \textit{go straight stop}, \textit{left turn stop}. Ground Truth (GT): \textit{go straight pass}, \textit{left turn pass}. (\textbf{b}) Interference from front tail lights and street lights. Predict: \textit{go straight stop}, \textit{left turn stop}. GT: \textit{go straight pass}, \textit{left turn pass}. }\label{fig:7}
\end{figure}

% We collected several day-time failure case scenes from two public datasets and exhibited them in Fig \ref{fig:6}. From Fig \ref{a} and Fig \ref{b} we can discover if the traffic light has been severely blocked by some large trucks or vans for a long time. In that case, our proposed algorithm will lose the clue from the traffic light signal to determine the passable status for the current intersection's available driving directions. As a result, the passable status generated from our model comes from random guesses. The larger time interval for each frame in the image buffer may be a remedy to mitigate this drawback.
% What's more, dramatic ambient light changes will also affect the proposed algorithm's performance. For example, as exhibited in Fig \ref{c} and Fig \ref{d}, if the ambient light is too dim or bright, and the ego vehicle is far away from the traffic light, the signal of the traffic light is too weak to be picked up by our model due to the severe interference from the ambient light environment. As a result, our proposed algorithm will fail to extract useful information and make a wrong judgment.

The typical failure cases are also collected for the night-time scene and exhibited in Fig. \ref{fig:7}. The reasons for the failure of the night scenes are very similar. Due to numerous light source interference, i.e. tail lights, street lights, headlights, and less texture information in the image, our model leverages the information from the wrong attention area. Therefore, it generates an incorrect prediction based on inaccurate feature information and ultimately fails. This problem is difficult to solve at the algorithm level since the input image has lost a lot of detailed information. Adding a post-processing voting scheme may suppress rapid state transitions and mitigate this drawback. 
 
\section{Conclusion and Future Work}\label{conclusion}
This paper introduced a novel end-to-end right-of-way status recognition algorithm for available driving directions at urban intersections. It takes multiple historical image frames as input and produces the right-of-way status for available driving directions as output. We proposed the TSA module to aggregate temporal information from the image buffer. In addition, the SCA module was proposed to enable our image embedding query to interact with the feature map. Finally, a multi-weight arcface class decoder was proposed to enhance the model's classification performance. The proposed algorithm is first trained and tested on the Bosch and Kaggle Lisa traffic light datasets to validate the performance. This is followed by several ablation studies on the Bosch and Kaggle Lisa datasets by removing the TSA module and by changing the number of cluster centres in the multi-weight arcface class decoder module to verify their importance in making the right-of-way status classification, and prove the superior performance of our model.

The right-of-way status recognition for available driving directions at urban signalised intersections is an important task related to driving safety. According to the failure case analysis presented in this paper, although our algorithm can accurately give the right-of-way status for each driving direction in the vicinity, it can fail when the traffic lights are far away from the ego vehicle, or during dramatic ambient light changes. For future research, we will consider incorporating map information and focus on longer-range classification.

\bibliographystyle{IEEEtran}
\bibliography{mybibtex}

\end{document}